\documentclass[preprint]{elsarticle}

\usepackage{mathrsfs}
\usepackage{amssymb}
\usepackage[cmex10]{amsmath}
\usepackage{amsthm}
\usepackage{graphics}
\usepackage{latexsym}
\usepackage{amsfonts}
\usepackage{lineno,xcolor}
\usepackage{IEEEtrantools}
\usepackage{subfigure}

\usepackage{color}
\usepackage{ctable}
\usepackage{graphicx}
\usepackage{verbatim}
\usepackage{epstopdf}
\usepackage{dsfont}

\newtheorem{theo}{Theorem}
\newtheorem{lemm}[theo]{Lemma}

\newtheorem{prop}[theo]{Proposition}
\newtheorem{defi}[theo]{Definition}

\theoremstyle{remark}
\newtheorem{exam}{Example}
\newtheorem{rema}[theo]{Remark}

\newcommand{\be}{\begin{IEEEeqnarray*}{rCl}}
\newcommand{\ee}{\end{IEEEeqnarray*}}
\newcommand{\ben}{\begin{IEEEeqnarray}{rCl}}
\newcommand{\een}{\end{IEEEeqnarray}}
\newcommand{\lb}[1]{\left[\begin{array}{#1}}
\newcommand{\rb}{\end{array}\right]}
\newcommand{\lp}[1]{\left(\begin{array}{#1}}
\newcommand{\rp}{\end{array}\right)}

\newcommand{\leftd}[1]{\left\{\begin{array}{#1}}
\newcommand{\rightd}{\end{array}\right.}

\def\A {\mathbf{A}}
\def\B {\mathbf{B}}
\def\C {\mathbf{C}}
\def\D {\mathbf{D}}

\def\I {\mathbf{I}}

\def\K {\mathbf{K}}
\def\L {\mathbf{L}}
\def\M {\mathbf{M}}

\def\P {\mathbf{P}}

\def\W {\mathbf{W}}

\def\a {\mathbf{a}}

\def\e {\mathbf{e}}
\def\f {\boldsymbol{f}}

\def\h {\boldsymbol{h}}

\def\s {\mathbf{s}}

\def\u {\mathbf{u}}
\def\v {\mathbf{v}}
\def\w {\mathbf{w}}
\def\x {\mathbf{x}}

\def\Bs {\mathscr{B}}

\def\Ac {\mathcal{A}}
\def\Bc {\mathcal{B}}

\def\Dc {\mathcal{D}}

\def\Ic {\mathcal{I}}
\def\Jc {\mathcal{J}}

\def\Nc {\mathcal{N}}
\def\Oc {\mathcal{O}}

\def\Sc {\mathcal{S}}

\def\xixi {\boldsymbol{\xi}}

\def\Lbd{\boldsymbol{\Lambda}}

\def\fii{\boldsymbol{\varphi}}

\def\Rb {\mathbb{R}}

\def\Eb {\mathbb{E}}

\def\Dds{\mathds{D}}
\def\Fds{\mathds{F}}

\def\Ods{\mathds{O}}
\def\Lds{\mathds{L}}
\def\TF{{\mathds{L}}}

\newcommand{\sign}{\mathrm{sign}}

\newcommand{\defeq}{\stackrel{\mathrm{def}}{=}}
\newcommand{\TR}{\mathsf{T}}

\newcommand{\cov}{\mathrm{Cov}}

\newcommand{\spanv}{\mathrm{span}}
\newcommand{\range}{\mathrm{range}}

\title{A study of the fixed points and spurious solutions of the FastICA algorithm}
\author[wtw]{Tianwen~Wei \corref{cor1}}
\address[wtw]{Zhongnan University of Economics and Law, Department of Statistics and Mathematics, Wuhan, China}
\cortext[cor1]{Corresponding author.  Email address: tianwen.wei.2014@ieee.org.\\
Part of this work \cite{WEISSP1} was presented  at IEEE workshop on Statistical Signal Processing 2014, Gold Coast, Australia.}
\begin{document}
\begin{frontmatter}

\begin{abstract}
The FastICA algorithm is one of the most popular iterative algorithms in the domain of linear independent component analysis. Despite its success, it is observed that FastICA occasionally yields outcomes that do not correspond to any true solutions (known as demixing vectors) of the ICA problem. These outcomes are commonly referred to as spurious solutions. Although FastICA is among the most extensively studied ICA algorithms, 
 the occurrence of spurious solutions are not yet completely understood by the community. 
 In this contribution, we aim at addressing this issue.
In the first part of this work, we are interested in the relationship between demixing vectors, local optimizers of the contrast function and (attractive or unattractive) fixed points of FastICA algorithm. Characterizations of these sets are given, and an inclusion relationship  is discovered.
In the second part, we investigate the possible scenarios where spurious solutions occur.
 We show that when certain bimodal Gaussian mixtures distributions are involved, there may exist spurious solutions that are attractive fixed points of FastICA. In this case, popular nonlinearities such as ``gauss'' or ``tanh'' tend to yield spurious solutions, whereas only ``kurtosis'' may give reliable results.
  Some advices are given for the practical choice of nonlinearity function.
  \end{abstract}

\end{frontmatter}
\section{Introduction}
{T}{he}  {Independent Component Analysis (ICA)} \cite{COMOBOOK, HYVABOOK},
is a statistical and computational method which aims at extracting the unobserved source signals from their linear mixtures without prior information on the
statistical properties of the unknown signals and on the mixing process.
As the name suggests, the fundamental assumption of ICA is that the source signals are statistically independent.
Up to date, there exist various ICA algorithms \cite{CARD1993, COMO94, HYVAOJA97, RADICAL, ZARZ2010} in the community, see \cite{CHEV04} for a survey. One of the most widely used ICA algorithms is the FastICA algorithm, proposed by Hyv\"arinen and Oja from the Finnish school \cite{HYVABOOK, HYVAOJA97,HYVA99}. It is based on the optimization of
a contrast function measuring the non-Gaussianity of the mixture, and it is derived as an approximation of Newton's method on the unit sphere.
The popularity of FastICA can be attributed to its simplicity, ease of implementation, and flexibility to choose the nonlinearity function.

There are two versions of FastICA algorithms: The one-unit (deflation) FastICA, and the symmetrical FastICA.
The one-unit version of FastICA corresponds to the sequential source separation scheme: it extracts one source at a time until all the sources are recovered, and to avoid that the algorithm converges to the same source twice, an additional deflationary procedure is required \cite{DELF}.
The one-unit FastICA has the common drawback of all sequential source separation scheme: the error propagation during successive extraction for problems with large dimensionality.
The symmetrical version of FastICA \cite{OJAYUAN} extract all the source signals simultaneously. It can be considered as several one-unit FastICA implemented in parallel, with the projection step replaced by an matrix orthonormalization in each iteration.
Symmetrical FastICA do not suffer the disadvantage of error propagation. However,  the downside of this version
is its unnecessary high computation load if only a small subset
of sources needs to be extracted from a high dimensional data
set.
This paper focuses only on the one-unit version of FastICA.

The FastICA algorithm has been extensively studied during the past years. It is shown to possess locally at least quadratic convergence speed, and in some cases, e.g. with ``kurtosis'' nonlinearity function, the convergence speed is even cubic \cite{HYVA99, SHEN}.
 Besides, it is also proved that the convergence of FastICA is monotonic \cite{REGA}. 
  The asymptotic performance of the algorithm is also investigated, first in \cite{HYVA97}, then in \cite{TICHOJA, OLLI}.
  It is worth mentioning that the Cramer-Rao bound of linear ICA is studied in \cite{TICHOJA}, where the authors show that if the nonlinearity function is adapted to the distributions of the source signals, then under some conditions the FastICA algorithm yields an asymptotically efficient estimator.
  The latest account of the asymptotic performance of FastICA is \cite{WEI2}, where the asymptotic covariance matrices are derived and compared under different scenarios depending on whether or not the centering and whitening procedure is exact.

In this work, we focus on studying the limit set (i.e. set of fixed points) of FastICA and investigating  occurrence of spurious solution.
 It is well-known that the limit of one-unit FastICA largely depends on the initial input of the algorithm, and it is in general not known to which vector FastICA converges. An vital question is whether or not the algorithm will always converges to a demixing vector. The answer to this question relies on the understanding of the relationship between the demixing vectors, the optimizers of the contrast function and the (attractive or unattractive) fixed points of FastICA algorithm. The first part of this paper will be devoted to the investigation of these sets. Some characterizations will be given and an inclusion relationship will be established.
 In particular, we show that FastICA algorithm based on ``kurtosis'' nonlinearity possess the desired property that only demixing vectors can be attractive fixed points. This latter result was first derived in \cite{DOUG2003}, where the proof was conducted in a different manner. The second part of the paper is devoted to the investigation of spurious solutions. Spurious solutions have already been noticed by some authors \cite{TICHOJA}, and were reported as ``saddle points'' of the contrast function. However, we show that this ``saddle point'' description is not accurate, because a spurious solution can very well be a global maximum or minimum point of the contrast function. In this work, we categorize spurious solutions as attractive or unattractive fixed points. We show that unattractive fixed points widely exist, and can potentially cause the phenomenon of ``false convergence'': the algorithm is
  considered ``converged'' and therefore halted by the stopping criterion before it actually reaches its true limit. It occurs when the initial iterate of the algorithm happens to locate in a small neighbourhood of a fixed point. We propose to adopt a strict stopping criterion in order to reduce the occurrence of this type of spurious solution. The second category of spurious solutions consists of spurious attractive fixed point.
  In most cases, attractive fixed points are desired demixing vectors, but this is not always true. Inspired by \cite{VRIN2005}, we test various bimodal distributions, and find that when some sources have certain asymmetrical bimodal distributions with Gaussian mixture, spurious attractive fixed point emerges even for commonly used nonlinearity functions such as ``Gauss'' and ``tanh''. In this case, ``kurtosis'' is the only reliable choice of nonlinearity. Finally, we discuss the impact of sampling error. Some advices are given with regards to the practical choice of nonlinearity.

 This paper is organized as follows. In Section II, we introduce all the basic notions of linear ICA: model, data preprocessing, contrast function and one-unit FastICA.
 Section III aims at characterizing four sets of interest, namely, the set of demixing vectors, the set of attractive fixed points, the set of local optimizers of the contrast function and the set of all (attractive and unattractive) fixed points.  In Section IV, we investigate the possible scenarios where spurious solutions occur and discuss the practical choice of nonlinearity function.
The concluding remarks of Section V bring the paper to an end.

\section{ICA data model and method}
In sequel, we use boldface uppercase letters such as $\M$ to denote matrices and 
boldface lowercase letters such as $\v$ to denote vectors.
 We denote by $\M^\TR$ the matrix transpose of $\M$ and $\|\M\|$ its spectral norm. With a slight abuse of notation,  $\|\cdot\|$ also stands for the Euclidean norm for vectors.
\subsection{ICA Data model}
We consider the following noiseless linear ICA model:
\ben
\x=\A\s,\label{ICAmodel1}
\een
where
\begin{enumerate}
\item $\s\defeq(s_1,\ldots, s_d)^\TR$ denotes the unknown \emph{source signals}. The components $s_1,\ldots, s_d$ are mutually independent, and at most one of them is Gaussian.
\item $\x\defeq(x_1,\ldots, x_d)^\TR$ denotes the \emph{observed signals}.
\item $\A\defeq (\a_1,\ldots, \a_d)$ is an unknown invertible square matrix, called the \emph{mixing matrix}.
\end{enumerate}
The task of ICA is to recover the source signal $\s$ based on the observation of $\x$ only, and this can be achieved by estimating the inverse of the mixing matrix $\A$. Note that since neither $\A$ nor $\s$ is known, the magnitude of $\s$ is not identifiable. To reduce this indeterminacy, we make the popular convention $\cov(\s)=\I$. Besides,
by centering and whitening the observed signal, i.e. setting $\tilde{\x}  =  \cov(\x)^{-\frac{1}{2}}(\x-\Eb[\x])$,
we can always transform model (\ref{ICAmodel1}) into an equivalent one:
\ben\label{ICAmodel2}
\tilde{\x}=\widetilde{\A}{\s},
\een
where $\tilde{\x}$ has zero mean and unit variance, and the new mixing matrix $\widetilde{\A}=(\A\A^\TR)^{-1/2}\A$ is orthogonal. Thus without loss of generality, we may as well directly suppose $\Eb[\s]=0$ and $\A$ is orthogonal in model (\ref{ICAmodel1}).

 It is well known \cite{COMO94} that under these assumptions,  we can only recover $\s$ up to a permutation and the sign. That is, it is only possible to find a matrix
 $\W^*=(\w^*_1,\ldots,\w^*_d)^\TR$ such that $\W^*\A=\Lbd\P$ where $\Lbd$ is a diagonal matrix with diagonal elements being $1$ or $-1$, and $\P$ is a permutation matrix. In the sequel, we call such $\W^*$
 the \emph{demixing matrix}, and rows of $\W^*$  the \emph{demixing vectors}. Clearly, a vector $\w^*$ can be a demixing vector if and only if there exists some $i\in\{1,\ldots,d\}$ such that $\w^*=\a_i$ or $-\a_i$.

\subsection{Contrast function}
In principle, two approaches can be adopted to estimate the demixing matrix $\W$: rows of $\W$ can be estimated either simultaneously, or sequentially.
The one-unit FastICA corresponds to the latter approach.
The estimation of rows of $\W$ is usually achieved by optimizing a criterion \cite{COMOBOOK,COMO94} called \emph{contrast} or \emph{contrast function}, that is a mapping $\Jc(\w)$ from $\Rb^d$ to $\Rb$ subject to the constraint $\|\w\|=1$. Contrast function can be considered as a measure of non-Gaussianity or independence, we refer to \cite{COMOBOOK, HYVABOOK} for more detail.
In this paper we consider the following type of contrast function:
\ben\label{contrastFunction}\label{contrast1}
\Jc(\w)=\Eb[G(\w^\TR\x)],\quad \w\in\Sc,
\een
where $G(\cdot):\Rb\to\Rb$ is a twice continuously differentiable nonlinear
and nonquadratic function\footnote{We implicitly require that the nonlinearity should be such that mathematical expectation (\ref{contrastFunction}) is well defined.} called the \emph{nonlinearity}
and $\Sc\defeq \{\w\in\Rb^d: \|\w\|=1\}$ stands for the unit sphere. In order to be consistent with the notation used in \cite{HYVAOJA2000, HYVA99}, we write $g(x)\defeq G'(x)$, the derivative of $G(x)$. When there is no risk of confusion, both $g(\cdot)$ and $G(\cdot)$ may be referred to as the nonlinearity function.
The choice of nonlinearity functions can be quite flexible. 
Popular nonlinearity  functions \cite{HYVA99} include the following: ``kurtosis'': $G_1(x)=x^4/4$; ``Gauss'': $G_2(x)=-\exp(-\frac{x^2}{2})$; ``tanh'': $G_3(x)=\log\cosh(x)$.
All of these nonlinearities are smooth, even, and can be bounded by a polynomial function.


It is shown in \cite{HYVA99} that contrast function having form of (\ref{contrastFunction}) can be utilized as a valid contrast for ICA in the sense that for $i=1,\ldots,d$, the vector $\pm\a_i$ is either a local minimizer or local maximizer of $\Jc$ provided that
\ben\label{condition1}
\Eb[g'(\pm s_i) \mp s_i g(\pm s_i)]\neq 0,
\een
where $g'$ denotes the derivative of $g$.
Note that if the signal $\s_i$ has a symmetrical distribution, or the nonlinearity $G$ is even, then we have
\be
\Eb[g'(s_i) -s_ig(s_i)]=\Eb[g'(-s_i) -(-s_i)g(-s_i)].
\ee
\subsection{FastICA algorithm}
The one-unit FastICA algorithm is an iterative method that searches the local optimizers of the contrast function ($\ref{contrastFunction}$). Using the following notation
\ben
\h(\w) & \defeq & \Eb[g'(\w^{\TR}\x)\w - g(\w^{\TR}\x)\x], \label{52} \\
\f(\w) & \defeq & \frac{\h(\w)}{\|\h(\w)\|}, \label{52a}
\een
we can describe the  FastICA algorithm as follows:
\begin{enumerate}
\item[1).] Choose  an arbitrary  initial iterate $\w^{(0)}\in\Sc$;
\item[2).] Run iteration $\w \leftarrow \f(\w)$ until convergence.
\end{enumerate}
In the sequel, we will refer to mapping (\ref{52a}) as the \emph{FastICA function}.
It is known \cite{HYVA99} that starting in a neighbourhood of $\pm\a_i$ for any $i$, the FastICA algorithm yields a sequence $\{\w^{(n)}\}$ that converges quadratically to $\pm\a_i$ if condition (\ref{condition1}) is met. We point out that under certain situations, FastICA oscillates
between neighborhoods of two antipodes on the unit sphere, which both provide the same correct separation (i.e. $\a_i$ and $-\a_i$). In this case, it is still considered that FastICA has successfully ``converged''.
In what follows,  we say $\{\w^{(n)}\}$ converges to $\v$ in strict sense if $\lim_{n\to\infty} \|\w^{(n)}-\v\|=0$, and in wide sense if
\ben\label{wideConvergence}
\lim_{n\to\infty} \inf \{\|\w^{(n)}-\v\|, \|\w^{(n)}+\v\|  \}=0.
\een


\subsection{Four sets}
Let us begin by defining several terms that will be used throughout this work.
\begin{defi}
\begin{itemize}
\item[-] An outcome $\u$ of the FastICA algorithm is a spurious solution if $\u\neq \pm\a_i$ for all $i=1,\ldots,d$.
 \item[-] A vector $\v$ is a fixed point of the FastICA function if $\f(\v)=\pm\v$.
\item[-] A fixed point $\v$ is called attractive if $\|\f'(\v)\|<1$, and unattractive if $\|\f'(\v)\|\geq 1$.
\end{itemize}
\end{defi}
Note that those points satisfying
$\f(\v)=-\v$ are not fixed points in strict sense, but we should still take them into consideration due to the aforementioned sign-flipping phenomenon.
These points will be sometimes referred to as \emph{generalized} fixed points.
A vector $\v$ being fixed point of $\f$ does not guarantee that FastICA will converge to it. In fact, an iterated function will converge to its fixed point only if this fixed point is also attractive. Attractive fixed points can be characterized by the value of the first order derivative of the underlying function at these points, as what we did in the definition.

We are interested in the relation among the following sets: \\
$\quad\Dds\defeq\{\pm\a_i,\,\, i=1,\ldots, d\}$; \\
$\quad\Fds\defeq\{\v\in\Rb^d:\,\,\f(\v)=\pm\v  \}$;\\
$\quad\Lds\defeq\{\v\in\Fds:\,\, \|\f'(\v)\|<1\}$;\\
$\quad\Ods\defeq\{\v\in\Sc:\, \v \textrm{ is a local optimizer of } \Jc\textrm{ on }\Sc\}$.\\
The meaning of these sets are obvious. The set $\Dds$ consists of the desired solutions (demixing vectors) of the ICA problem;
 $\Fds$ is the set of all fixed points of the FastICA function; $\Lds$  is the set of attractive fixed points; $\Ods$ stands for the set of local minimizers and maximizers of the contrast function $\Jc$.

A vital question is to which set does FastICA converge. We hope the algorithm will converge to $\Dds$, since this would give the correct solution of our ICA problem. However, classical results \cite{HYVA99} only confirms that $\Dds\subset \Ods$ and $\Dds\subset\Lds$ provided that (\ref{condition1}) holds,
and it is not known  if these inclusions are strict\footnote{In this paper, notation $\subset$ stands for the ``subset''  rather than the ``proper subset'' inclusion. Hence $\Dds\subset \Lds$ does not exclude $\Dds=\Lds$.}. Besides, the relationship between $\Ods$ and $\Lds$ is still unclear.
In the next section, we will give a complete characterization of these sets.

\section{Demixing vectors of ICA and fixed points of FastICA algorithm}
\subsection{Assumptions \label{assumption12}}
In the sequel, we make the following assumption:
\begin{IEEEeqnarray}{lllll}
\Ac_1:\quad& G(x)   &=   &G(-x), \quad\quad&\forall x\in\Rb; \\
\Ac_2:\quad&\h(\w)  &\neq& 0,    \quad\quad&\forall\w\in\Sc^d.
\end{IEEEeqnarray}
Assumption $\Ac_1$ is very popular in the community of ICA.
One major advantage of even nonlinearity is that it enables a cubic convergence speed of FastICA \cite{SHEN}, provided that the corresponding source signal has a symmetrical distribution. Here, this assumption mainly serves to simplify the convergence analysis in the sign-flipping case. In fact, if the underlying nonlinearity is not even, then $\f(\a_i)=-\a_i$ does not necessarily imply $\f(-\a_i)=\a_i$, which potentially makes the algorithm less trackable.
Assumption $\Ac_2$ is made to guarantee that the FastICA function is well-defined everywhere in $\Sc$ so that the algorithm will not suddenly stop due to the occurrence of infinity. Note that condition (\ref{condition1}) is a corollary of $\Ac_2$, as will be subsequently pointed out.

 \subsection{Characterization of $\Fds$}
 Observe that for any input $\v\in\Rb^d$,  the projection step in (\ref{52a}) does not change the direction of $\h(\v)$. Thus, a vector $\v$ can be a fixed point of $\f$ if and only if $\v\in\Sc$ and it is parallel to $\h(\v)$, i.e. there exists $\alpha(\v)\neq 0$ such that by (\ref{52})
\ben\label{54a}
\h(\v)&=&\Eb[g'(\v^{\TR}\x)\v - g(\v^{\TR}\x)\x]=\alpha(\v)\v.
\een
Using the orthogonal decomposition
\ben\label{decomposition}
\x=(\I-\v\v^\TR)\x + (\v\v^\TR)\x,
\een
we can write $\h(\v)$ as
\be
\h(\v)&=&\Eb[g'(\v^{\TR}\x) - g(\v^{\TR}\x)\v^\TR\x]\v + \Eb[g(\v^{\TR}\x)(\I-\v\v^\TR)\x].
\ee
Note that the second term on the right hand side is perpendicular to $\v$, thus (\ref{54a}) holds if and only if
\ben
\alpha(\v)&\defeq& \Eb[g'(\v^{\TR}\x) - g(\v^{\TR}\x)\v^\TR\x]  \neq   0, \label{defAlpha}\\
\fii(\v) &\defeq & (\I-\v\v^\TR)\Eb[g(\v^{\TR}\x)\x] =0. \label{defFii}
\een
Since $\h(\v)\neq 0$ by assumption $\Ac_2$, $\fii(\v)$ and $\alpha(\v)$ cannot be both zero. This means that condition (\ref{defFii})  alone can be used to characterize $\Fds$. We state this result in the following Lemma:
\begin{lemm}\label{lemmaFds}
$\Fds=\{\v\in\Sc:\,\, \fii(\v)=0  \}$.
\end{lemm}
\begin{rema}
From the assumption that $G$ is even, we deduce immediately from (\ref{defAlpha}) and (\ref{defFii}) that $\alpha(\cdot)$ is also even while $\fii$ is odd. Hence, if $\v\in\Fds$, then we have
$\alpha(-\v)=\alpha(\v)\neq 0$ and $\fii(-\v)=-\fii(\v)=0$. This means $-\v\in\Fds$ as well. Besides, we have
 \be
 \f(\v)=\frac{\h(\v)}{\|\h(\v)\|} = \frac{\alpha(\v)\v}{|\alpha(\v)|}=\mathrm{sign}(\alpha(\v))\v.
 \ee
It follows that if $\alpha(\v)>0$, then it is a strict fixed point, otherwise it is a generalized one. The same argument also applies to $-\v$.
 Note that if $\alpha(\v)=\alpha(-\v)<0$,
 then the sign-flipping phenomenon occurs: self-iteration of $\f$ at $\v$ yields a sequence of flipping signs: $\v,-\v,\v, -\v,\ldots$.
\end{rema}
From Lemma \ref{lemmaFds}, we then deduce the following inclusion: 
\begin{lemm}\label{55c}
$\Dds\subset\Fds$.
\end{lemm}
\begin{IEEEproof}
Let us consider $\v=\a_i\in\Dds$ for some $i\in\{1,\ldots,d\}$ and the decomposition (\ref{decomposition}).  Since $\a_i\a_i^\TR\x=\a_i s_i$
and $(\I-\a_i\a_i^\TR)\x=\x-\a_is_i=\sum_{j\neq i}^d\a_js_j$ by ICA model (\ref{ICAmodel1}),
it follows that $\a_i^\TR\x$ and $(\I-\a_i\a_i^\TR)\x$ are independent.
As a result, we have
\be
\Eb[g(\a_i^{\TR}\x)(\I-\a_i\a_i^\TR)\x]=\Eb[g(\a_i^{\TR}\x)]\Eb[(\I-\a_i\a_i^\TR)\x]=0.
\ee 
\end{IEEEproof}

\subsection{Characterization of $\Lds$}
The characterization of $\Lds$ involves the first-order derivative of the FastICA function. Direct derivation of (\ref{52}) and (\ref{52a}) yields \cite{WEI2}:
\ben
\h'(\w)
& = & \Eb[g''(\w^{\TR}\x)\w\x^{\TR}  + g'(\w^{\TR}\x)\I - g'(\w^{\TR}\x)\x\x^{\TR}], \nonumber\\
\IEEEstrut[8\jot]
\f'(\w) & = & \frac{\big(\|\h(\w)\|^2\I-\h(\w)\h(\w)^{\TR}\big) \h'(\w)}{\|\h(\w)\|^3}. \label{df}
\een
We can show that if $\v\in\Fds$, then the condition $\|\f'(\v)\|<1$ is equivalent to
\ben
\|(\I-\v\v^\TR)\Eb[g'(\v^{\TR}\x)(\I-\x\x^{\TR})]\|< |\alpha(\v)|. \label{LemmaTF}
\een
We state this characterization formally in the following lemma.
\begin{lemm}\label{55a}
$\Lds=\{\v\in\Fds:\,\, \v\,\,\, \mathrm{verifies }\,\,(\ref{LemmaTF}).\}$
\end{lemm}
\begin{IEEEproof}
Recall that if $\v\in\Fds$ then
$\h(\v)=\alpha(\v)\v$.
It follows that
\be
\|\h(\v)\|^2\I-\h(\v)\h(\v)^{\TR}
& = & \alpha(\v)^2(\I-\v\v^\TR).
\ee
Then the numerator of (\ref{df}) becomes
\be
&&\alpha(\v)^2(\I-\v\v^\TR) \Eb[g''(\v^{\TR}\x)\v\x^{\TR}  + g'(\v^{\TR}\x)\I - g'(\v^{\TR}\x)\x\x^{\TR}]\\
& =& \alpha(\v)^2 \Eb[g'(\v^{\TR}\x)(\I -\v\v^\TR) - g'(\v^{\TR}\x)(\I-\v\v^\TR)\x\x^{\TR}] \\
& =& \alpha(\v)^2 (\I-\v\v^\TR)\Eb[g'(\v^{\TR}\x)(\I-\x\x^{\TR})].
\ee
As a result,
\ben\label{fprime}
\f'(\v)=\frac{(\I-\v\v^\TR)\Eb[g'(\v^{\TR}\x)(\I-\x\x^{\TR})]}{ |\alpha(\v) |}.
\een
From (\ref{fprime}), we conclude that
$\|\f'(\v)\|<1$ if and only if (\ref{LemmaTF}) holds.
\end{IEEEproof}
\begin{rema}
Since $G$ is even, both $g'$ and $\alpha$ are even functions. It then follows from (\ref{fprime}) that $\f'(\v)=\f'(-\v)$. Thus $\v\in\Lds$ if and only if $-\v\in\Lds$.
\end{rema}
\begin{lemm}\label{DinTF}
$\Dds\subset\Lds$.
\end{lemm}
\begin{IEEEproof}
If $\v\in\Dds$, then (\ref{fprime}) holds since $\Dds\subset\Fds$. Suppose $\v=\a_i$ for some index $i$.  As is shown in the proof of Lemma \ref{55c},
$\a_i^\TR\x$ and $(\I-\a_i\a_i^\TR)\x$ are independent. From this and in view of (\ref{fprime}), we deduce immediately $\f'(\a_i)=0$.
 Then by the definition of $\Lds$, the desired inclusion follows.
\end{IEEEproof}
Next, we show that attractive fixed points are really attractive, that is, FastICA tends to converge to these points.
\begin{lemm}
For any $\v\in\Lds$, there exists a neighbourhood $\Bs_r(\v)$ such that for any $\w^{(0)}\in\Bs_r(\v)$, the FastICA algorithm converges to $\v$ in the sense of (\ref{wideConvergence}).
\end{lemm}
\begin{IEEEproof}
For $\v\in\Lds$ verifying $\f(\v)=\v$, the convergence can be easily proved \cite{WEI2} by a traditional fixed point argument. It suffices to
notice that
there exists $r>0$ such that
\be
\sup_{\w\in\Bs_r(\v)} \|\f'(\w)\|\leq K<1,
\ee
by the continuity of $\f'$.
Then for $\w^{(0)}\in\Bs_r(\v)\cap\Sc$, we have $\|\w^{(1)}-\v\|=\|\f(\w^{(0)})-\f(\v)\|\leq K\|\w^{(0)}-\v\|$. By induction, we can get
$\|\w^{(n)}-\v\|\leq K^nr$, and the convergence follows.

As for $\v\in\Lds$ such that $\f(\v)=-\v$, let us consider the \emph{change of nonlinearity}, i.e. we consider $\f^{-}$ that are defined as in
(\ref{52}) and (\ref{52a}) but with the underlying nonlinearity $G$ being replaced by $-G$. It is easy to verify $\f=-\f^-$, hence
 $\f^-(\v)=\v$ and $\|(\f^-)'(\v)\|=\|\f'(\v)\|<1$. Applying the previous result, we assert that there exists a neighbourhood $\Bs_r(\v)$ such that as long as the starting point $\w^{(0)}$ lies within, the FastICA algorithm using $\f^-$ converges to $\v$.
Let us denote by  ``$\circ$'' the function composition, i.e.
\be
\f\circ\f(\w)\defeq \f\big(\f(\w) \big).
\ee
Since $\h$ and $\f$ are odd, we have $\f^-(\w^{(0)})=-\f(\w^{(0)})$, $\f^-\circ\f^-(\w^{(0)})=\f^-\big( -\f(\w^{(0)})\big) = \f\circ\f(\w^{(0)})$, and more generally
 \ben \label{2013118}
(-1)^n\underbrace{\f\circ\cdots\circ\f}_n(\w^{(0)})=\underbrace{\f^-\circ\cdots\circ\f^-}_n(\w^{(0)}).
 \een
Note that the term on the right hand side of (\ref{2013118}) converges to $\v$ as $n$ tends to $\infty$. Therefore, we have
 \be
\lim_{n\to\infty} \inf \{\|\w^{(n)}-\v\|, \|\w^{(n)}+\v\|  \}=0,
\ee
with $\{\w^{(n)}\}$ being generated by $\f$.
\end{IEEEproof}
\begin{rema} If $\f'(\v)=0$, which is the case for $\v\in\Dds$, then the FastICA algorithm converges locally with at least a quadratic convergence speed. For a more detailed account about the convergence speed of FastICA, we refer to \cite{SHEN,WEI2}.
\end{rema}

\subsection{Characterization of $\Ods$}
The property of $\Ods$ is investigated in detail in \cite{WEI2}. Here, we cite the following proposition therein:
\begin{prop}\label{secondapproxi}
For any $\w,\v\in \Sc$, we have
\ben\label{reprePhi}
\Jc(\w)
& =& \Jc(\v)+(\w-\v)^{\TR}\fii(\v) +\frac{1}{2}(\w-\v)^{\TR}\K(\v)(\w-\v) \nonumber \\
&& + \Oc(\|\w-\v\|^3),
\een
where $\fii(\v)$ is defined in (\ref{defFii}) and $\K(\v)$ is given by
\ben
\K(\v) & \defeq & \alpha(\v)\I + \L(\v), \label{1}\\
\L(\v)& \defeq & (\I-\v\v^\TR)\Eb[g'(\v^\TR\x)(\x\x^\TR-\I)](\I-\v\v^\TR) .\label{2}
\een
\end{prop} 
We emphasize that the representation of $\Jc$ given in (\ref{reprePhi}) holds only for $\w,\v\in\Sc$. Hence it is not an ordinary corollary of Taylor's Theorem.

The advantage of writing $\Jc$ in form of (\ref{reprePhi}) is that it reveals the necessary condition $\fii(\v)=0$ for $\v$ to be a local optimizer of $\Jc$. Since the condition $\fii(\v)=0$ defines the set of fixed points $\Fds$ by Lemma \ref{lemmaFds},  the following inclusion follows:
 \begin{lemm}\label{MinF}
 $\Ods\subset\Fds$.
 \end{lemm}
We can deduce from (\ref{reprePhi}) that if $\fii(\v)=0$ and the matrix $\K(\v)$ is either positive definite or negative definite,
then $\v$ is a local optimizer of $\Jc$. Next, we show that for $\v\in\Lds$ this condition is satisfied.

\begin{lemm}\label{TFinM}
$\Lds\subset\Ods$.
\end{lemm}
\begin{IEEEproof}
Let us write
\ben\label{matrixB}
\B(\v)\defeq (\I-\v\v^\TR)\Eb[g'(\v^{\TR}\x)(\I-\x\x^{\TR})],
 \een
 and denote respectively by $\lambda_{\min}(\cdot)$ and $\lambda_{\max}(\cdot)$ the smallest and the largest singular value of the underlying matrix. Recall that the singular values of a matrix $\M$ are defined as the square root of the eigenvalues of $\M^\TR\M$.

By (\ref{matrixB}), we have
\be
\L(\v) &=&-\B(\v)(\I-\v\v^\TR),\quad \v\in\Sc    \\
\f'(\v)&=&\frac{\B(\v)}{|\alpha(\v)|},\quad \v\in\Fds
\ee
 where the second equation is due to (\ref{fprime}). For $\v\in\Lds$, by Lemma \ref{55a} we have $\|\B(\v)\|< |\alpha(\v)|$. Since spectral norm is submultiplicative, there holds
\ben\label{inequality1}
\|\L(\v)\|\leq \|\B(\v)\| \|\I-\v\v^\TR\|<|\alpha(\v)|\cdot \|\I-\v\v^\TR\|.
\een
Note that $\I-\v\v^\TR$ is a projection matrix, hence its eigenvalues are either 0 or 1. Therefore $\|\I-\v\v^\TR\|\leq 1$ and inequality (\ref{inequality1})
becomes
\ben\label{118b}
\|\L(\v)\|< |\alpha(\v)|.
\een
Besides, since the matrix $\L(\v)$ is symmetrical, its singular values
coincide with the absolute value of its eigenvalues. Applying this result to (\ref{118b}) gives
\ben\label{55b}
-|\alpha(\v)| \leq \lambda_{\min}(\L(\v)) \leq \lambda_{\max}(\L(\v)) \leq |\alpha(\v)|.
\een
Combining (\ref{55b}) and (\ref{1}), we deduce that  $\K(\v)$ is positive definite if $\alpha(\v)>0$, and negative definite if $\alpha(\v)<0$. The case $\alpha(\v)=0$ is excluded since otherwise we would have $\h(\v)=0$, which contradicts assumption $\Ac_2$.
\end{IEEEproof}
\begin{rema}\label{517a}
 If $\v\in\Dds$, then we have actually $\L(\v)=0$. In this case, the matrix $\K(\v)$ is positive definite if and only if $\alpha(\v)>0$, and negative definite if and only if $\alpha(\v)<0$. Therefore $\pm\a_i$ is a local minimizer of $\Jc$ if $\alpha(\pm\a_i)>0$ and local maximizer if $\alpha(\pm\a_i)<0$.
 In particular, for ``kurtosis'' nonlinearity we have
 \be
 \alpha(\a_i)=\Eb[3(\a_i^{\TR}\x)^2 - (\a_i^{\TR}\x)^4]=3-\Eb[s_i^4] \defeq - \kappa_i,
 \ee
 where $\kappa_i$ denotes the fourth-order cumulant of $s_i$. If $s_i$ is sub-Gaussian, i.e. $\kappa_i<0$, then $\alpha(\a_i)>0$, hence $\a_i$ is a local minimizer of the contrast function. Likewise, if $s_i$ is super-Gaussian, then $\alpha(\a_i)<0$, which implies that $\a_i$ is a local maximizer.
\end{rema}
Combining Lemma \ref{DinTF}, Lemma \ref{MinF} and Lemma \ref{TFinM} together, we get the main result of this section:
\begin{theo}\label{main1}
$\Dds\subset\Lds\subset\Ods\subset\Fds$.
\end{theo}
One may ask if any of these inclusions is actually an equality. The answer is, in the general case, none of them are. Nevertheless, we have the following result
for ``kurtosis'' nonlinearity:
\begin{theo}\label{kurtosis}
For kurtosis nonlinearity function,
we have $\Dds=\Lds$. Moreover, for any $\v\in\Fds\backslash\Dds$, there holds $\|\f'(\v)\|=3$.
\end{theo}
\begin{IEEEproof}
See Appendix \ref{proofKurtosis}. The proof is based on \cite{DOUG2003}.
\end{IEEEproof}

\subsection{Practical situation}
In practice, we have only a finite and possibly noised sample of the observed signal $\x$ issued from model (\ref{ICAmodel1}):
\be
\x(t)=\A\s(t) + \xixi(t),\quad t=1,\ldots, N,
\ee
where $\xixi(t)$ is i.i.d. sequence of Gaussian noises with zero mean.

 We define the empirically centered and whitened data as
\ben\label{514e}
\tilde{\x}(t)\defeq \C_N^{-1/2}(\x(t) -\bar{\x} ),\,\, t=1,\ldots, N,
\een
where $\bar{\x}$ is the sample mean of $\x(t)$ and $\C_N$ is the empirical covariance matrix
\be
\C_N\defeq \frac{1}{N}\sum_{t=1}^N\big(\x(t) -\bar{\x}\big)\big(\x(t)-\bar{\x}  \big)^\TR.
\ee
The \emph{empirical} FastICA function $\widehat{\f}$ is defined as follows \cite{WEI2}:
\ben
\widehat{\h}(\w) & \defeq & \frac{1}{N}\sum_{t=1}^N\Big(g'(\w^{\TR}\tilde{\x}(t))\w - g(\w^{\TR}\tilde{\x}(t))\x(t)\Big), \label{521c} \\
\widehat{\f}(\w) & \defeq & \frac{\widehat{\h}(\w)}{\|\widehat{\h}(\w)\|}. \label{521d}
\een
The empirical one-unit FastICA algorithm \cite{HYVA99} is then simply the scheme of self-iteration $\w\leftarrow \widehat{\f}(\w)$. Introduce the empirical contrast function $\widehat{\Jc}$:
\ben\label{empiricalContrast}
\widehat{\Jc}(\w)=\frac{1}{N}\sum_{t=1}^NG(\w^\TR\tilde{\x}(t)),\quad \w\in\Sc.
\een

It has been shown in \cite{WEI2} with the assumption of the absence of noise that, starting in a neighbourhood of $\a_i\in\Dds$, the empirical FastICA algorithm with probability one for large enough $N$. Moreover, the limit, denoted by $\hat{\a}_i$, is independent of the starting position and is a consistent estimator of $\a_i$.
We refer the readers to \cite{WEI2} for a more detailed account of this matter.

For a given ICA model, the estimator $\hat{\a}_i$ depends only on the underlying nonlinearity function used in the algorithm.
The asymptotic variance of $\hat{\a}_i$, which is a measure the separation performance of the algorithm, is therefore determined solely by the nonlinearity, too.
The estimating problem, i.e. the problem of finding the optimal nonlinearity that achieves the efficiency, has already been studied \cite{TICHOJA, TICHOJA2}.

 Here, we are only interested in establishing an analogy of Theorem \ref{main1} for the empirical case.
  Denote\\
$\widehat{\Dds}\defeq\{\pm\hat{\a}_i,\,\, i=1,\ldots, d\}$; \\
$\widehat{\Fds}\defeq\{\v\in\Rb^d:\,\,\widehat{\f}(\v)=\pm\v  \}$;\\
$\widehat{\Lds}\defeq\{\v\in\widehat{\Fds}:\,\, \|\widehat{\f}'(\v)\|<1\}$;\\
$\widehat{\Ods}\defeq\{\v\in\Sc:\, \v \textrm{ is a local optimizer of } \widehat{\Jc} \textrm{ on }\Sc\} $.

We can show that an empirical version of Theorem \ref{main1} holds:
\begin{theo}
If the noise is absent, then we have
$\widehat{\Dds}\subset\widehat{\Lds}\subset\widehat{\Ods}\subset\widehat{\Fds}$ with probability one for large enough $N$.
\end{theo}
\begin{IEEEproof}
The inclusion $\widehat{\Dds}\subset\widehat{\Lds}$ holds trivially \cite{WEI2}, while
the proof of Lemma \ref{MinF} and Lemma \ref{TFinM} applies for inclusion $\widehat{\Lds}\subset\widehat{\Ods}\subset\widehat{\Fds}$ as well.
\end{IEEEproof}

\section{Investigation of spurious solutions of FastICA}
\subsection{General remark}
Iterative ICA algorithm such as FastICA may yield solution that does not correspond to the extraction of any independent component.
Such solution is called a \emph{spurious solution}.
If we are to have confidence in our ICA  algorithm,
then we should have a clear idea about
how often and under which circumstances these solutions may occur, and  if possible, take measures to reduce their occurrence.

Many factors may have an influence on the occurrence of spurious solutions. These factors include the distributions of the sources, the nonlinearity function, the initial iterate and the stopping criterion. In this section, we will study the impact of all these factors, with a focus on the choice of nonlinearity and choice of stopping criterion.

Let us begin by examining the nature of spurious solution.
 For contrast function having the form of (\ref{contrast1}),  it is already knew that there exist local optimizers that are not demixing vectors, i.e. $\Ods\backslash\Dds\neq \emptyset$. Theorem \ref{main1} tells us that all of these points do not cause trouble, because unlike ordinary gradient-based methods that search all the optimum points, the FastICA algorithm has the ability to filter a large proportion of spurious optimum points: it
may only get stuck at fixed points of the FastICA function (attractive or unattractive) and it
only converges to its attractive fixed points. This fact suggests us to focus our analysis on the attractive/unattractive fixed points of the algorithm, rather than the traditional subjects in the optimization theory, e.g. local optima/global optima/saddle points.

The following result says that spurious fixed points (i.e. points belonging to the set $\Fds\backslash\Dds$) widely exist on the unit sphere.
\begin{prop}\label{main2}
Let $\a_i,\a_j\in\Dds$ such that $\alpha(\a_i)$ and $\alpha(\a_j)$ have the same sign. Then there exists $0< c < 1$ such that $\v\defeq c\,\a_i + \sqrt{1-c^2}\a_j$ belongs to $\Fds$. Moreover, if the corresponding source signals $s_i$ and $s_j$ have identical distribution, then $\v= (\a_i + \a_j)/\sqrt{2}$.
\end{prop}
\begin{IEEEproof}
See Appendix \ref{proofmain2}.
\end{IEEEproof}
A spurious fixed point can be either attractive or unattractive. The following result reveals that for given nonlinearity function and source distributions, if spurious attractive fixed point (i.e. points belonging to the set $\Lds\backslash\Dds$) exists in 2-dimensional case, then it will also exist in higher dimensional case.
\begin{prop}\label{118c}
Let $G$ be a fixed nonlinearity function. Suppose that  there exists probability distributions $\Dc_1$, $\Dc_2$, such that in 2-dimensional case
 FastICA based on $G$ has a spurious attractive fixed point
 for $s_1\sim\Dc_1$ and $s_2\sim\Dc_2$.
 Then in the general $n-$dimensional case, if there exist two source signal $s_i,s_j$ with $i\neq j$ such that $s_i\sim\Dc_1$ and $s_j\sim\Dc_2$, then FastICA based on $G$ also has a spurious attractive fixed point.
\end{prop}
\begin{IEEEproof}
See Appendix \ref{ProofSpurious}.
\end{IEEEproof}

\subsection{Spurious solutions as unattractive fixed points}
Theoretically, unattractive fixed points should not be problematic since they are not ``stable'': the output of FastICA tends to move away from them unless the input of the algorithm is exactly the fixed point itself, which is a zero probability event. However, there is an algorithmic issue here: as we shall show, when certain stopping criterion is involved,
there exists the risk of false convergence: the algorithm is considered ``converged'' and therefore halted by the stopping criterion before it actually converges its true limit. It occurs when the initialization of the algorithm happens to locate in a small neighbourhood of an unattractive fixed point.
To see this, let us examine the widely used stopping criterion  for one-unit FastICA proposed in \cite{TICHOJA}:
\ben\label{stopping}
1-|\w^{\TR}_{\mathrm{new}}\w_{\mathrm{old}}|<\epsilon
\een
where $\epsilon$ is a suitable constant and $\w_{\mathrm{old}}$ and $\w_{\mathrm{new}}$ are the outputs of FastICA in two consecutive iterations. If the initialization $\w^{(0)}$ locates sufficiently close to a fixed point $\v$, then
(\ref{stopping}) is satisfied and the algorithm will be immediately halted.
In fact,
 we have
\be
\f(\w^{(0)}) - \v=\f(\w^{(0)}) - \f(\v)=\f'(\xixi)(\w^{(0)} -\v )
\ee
where $\xixi\in\Bs_r(\v)$. Using triangular inequality, it is not difficult  to see that
 \be
 1- |\w^{(0)\TR}\w^{(1)}| 
& \leq & \frac{1}{2} (1+ \|\f'(\xixi)\|)^2 \|\w^{(0)} -\v \|^2.
 \ee
 If the initial iterate $\w^{(0)}$ is close enough to $\v$ such that
 \be
 \frac{1}{2} (1+ \|\f'(\xixi)\|)^2 \|\w^{(0)} -\v \|^2\leq \epsilon,
 \ee or equivalently
  \ben\label{616a}
  \|\w^{(0)} -\v \| \leq \frac{\sqrt{2\epsilon}}{1+\|\f'(\xixi)\|} \approx \frac{\sqrt{2\epsilon}}{1+\|\f'(\v)\|},
  \een
  then the algorithm will be considered converged by the stopping criterion after just one single iteration.


Several methods are proposed to alleviate the issue of false convergence.
The simplest way is to tighten the error tolerance $\epsilon$ used in the stopping rule.
Note that according to (\ref{616a}), the rate of false convergence is of order $\Oc(\epsilon^{1/2})$, which means to reduce the rate of false convergence by 10 times, one would need approximately an error tolerance 100 times smaller.
An alternative method is to impose a fixed minimum number of iterations, say $n=10$ in addition to (\ref{stopping}), so that
in case the initial iterate happens to locate near some spurious fixed point,
the algorithm would still be able to move away from that point before it is halted.
The third approach is the ``check of saddle points'' method proposed in \cite{TICHOJA}. It is essentially based on the observation that spurious solutions tend to occur near $(\a_i \pm \a_j)/\sqrt{2}$ for some $i,j$ (which is confirmed theoretically by Proposition \ref{main2}), and under the orthogonality constraint, they occur as a pair: if there is one spurious solution near $(\a_i + \a_j)/\sqrt{2}$, then there will be another one near $(\a_i - \a_j)/\sqrt{2}$.
It is then proposed to run a comparative test
of non-Gaussianity for all the estimates $\hat{\a}_1,\ldots, \hat{\a}_d$ along with $(\hat{\a}_i \pm \hat{\a}_j)/\sqrt{2}$ for all pairs of $(i,j)$.
If the pair $(\hat{\a}_i + \hat{\a}_j)/\sqrt{2}$ and $(\hat{\a}_i - \hat{\a}_j)/\sqrt{2}$ yields a larger non-Gaussianity index than $\hat{\a}_i$ and $\hat{\a}_j$, then the former pair will be accepted as correct estimates of demixing vectors whereas the latter will be rejected as spurious solutions.
Lastly, there is the ICASSO method \cite{HIMB}, which is based on running FastICA several times with different initial iterates and (or) resampling of the data. 
The idea is that a tight cluster of estimates is considered to be a candidate for including a ``good'' estimate and a centroid of such cluster is considered a more reliable estimate than any estimate from an arbitrary run.
 A thorough comparative study of the methods introduced above is beyond the scope of this work.

\begin{exam}
In Table \ref{table1}, we investigate the occurrence of spurious solutions under four different stopping criteria, namely
 $\epsilon=10^{-4},10^{-6},10^{-8}$ along with a combination of $\epsilon=10^{-4}$ and a minimum iteration number $n=10$. Various scenarios are tested: all three popular nonlinearites ``kurtosis'', ``Gauss'' and ``tanh'', various source distributions and three different dimensionalities $d=2$, $d=3$ and $d=5$.
The theoretical value of $\|\f'(\v)\|$,
where $\v=(\a_i+\a_j)/\sqrt{2}$ is the potential spurious solution as a fixed point (see Proposition \ref{main2}), is also marked in each scenario as a reference.
All sources are set to be identically distributed so that the message conveyed by the simulation results can be clear.
The sample size is fixed at $N=5000$, large enough so that the finiteness of sample size has almost no impact on the occurrence of spurious solutions: if a spurious solution eventually occur, then almost surely it is not introduced by the sampling error.

There are three sub-tables.  Each sub-table corresponds to a different dimensionality and is divided into two parts according to the distribution of the sources. In the upper part, all distributions involved are symmetrical: ``sinus'' means the distribution of $\sqrt{2}\sin(x)$, where $x$ is
uniformly distributed in $(0, 2\pi)$ ,  ``GG($\alpha$)''
means the generalized Gaussian distribution with parameter $\alpha$,  ``Bimod($\mu$)'' stands for the symmetrical bimodal distribution with Gaussian mixture, and ``bpsk'' is the discrete distribution with equiprobable values $\pm1$.
In the lower part of the table, ``Bimod($\mu_1$,$\mu_2$)'' stands for the asymmetrical bimodal distributions with two modes $\mu_1$ and $\mu_2$.
Both  ``GG'' and ``Bimod'' families will be described in Appendix \ref{PDF}.
During the entire simulation, we used a relatively large sample size $N=5000$, so that the sampling error is not influential.

From Table \ref{table1}, we observe that in all three scenarios a lower false convergence rate always comes with a larger $\|\f'(\v)\|$ value, which is expected (see e.g. (\ref{616a})). Besides, we notice that the value of $\|\f'(\v)\|$ is independent of the model dimensionality in the case of i.i.d. sources, as can be deduced from the proof of Proposition \ref{118c}.
It is shown in the tables that simply imposing a much lower error tolerance $\epsilon=10^{-8}$ or a minimum iteration number such as $n=10$ does significantly reduce the occurrence of false convergence.
We also observe that when the asymmetrical bimodal distributions are involved, for nonlinearities such as ``Gauss'' and ``tanh'', the spurious solutions systematically occur and cannot be reduced merely by adopting a stricter stopping criterion. This is because these spurious solutions have a different nature: they are attractive fixed points of the FastICA algorithm, corresponding to a $\|\f'\|$ value strictly smaller than 1. We discuss this category of spurious solutions separately in the next section.
\end{exam}

\subsection{Spurious solutions as attractive fixed points}
\begin{figure*}[!t]
\centerline{
\subfigure[]{\includegraphics[width=4.25in]{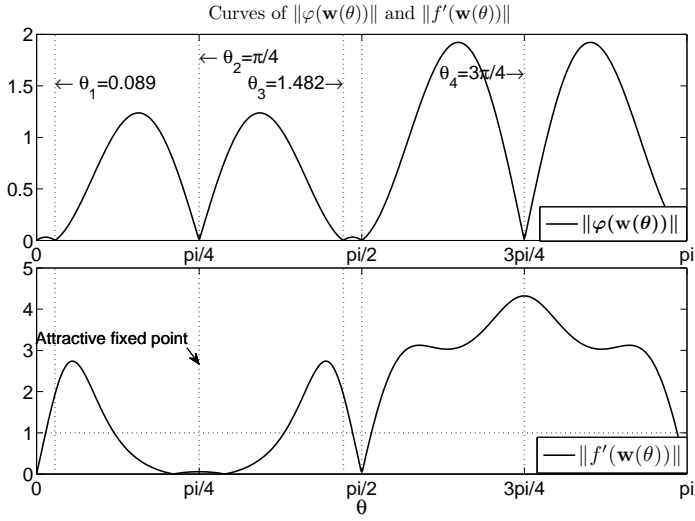}
\label{wei1}}\hspace{-3em}
\subfigure[]{\includegraphics[width=4.1in]{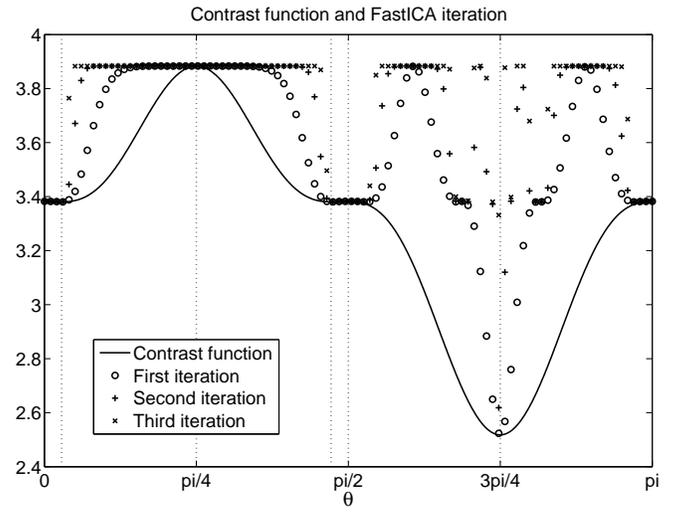}
\label{wei2}}}
\caption{ The presence of attractive and non attractive fixed points that are not demixing vectors. Two source signals $s_1,s_2\sim\,$Bimod(-0.4, 2), and $g(x)=x^5$. }
\end{figure*}

We have seen that for ``Gauss'' and ``tanh'' nonlinearity functions, there may exist spurious fixed points that are attractive; in other words, the inclusion $\Dds\subset\Lds$ is strict. Unlike the issue of false convergence which 
is not an intrinsic problem and indeed rarely happens, the spurious solutions as attractive fixed points ($\Lds\backslash\Dds$), if exist, will be generated by the algorithm  with very noticeable probability due to their large convergence domain.  These spurious solutions  cannot be eliminated or reduced by merely tightening the stopping criterion.
They may pass tests of non-Gaussianity (e.g. test used in the ``check of saddle points'' method, see \cite{TICHOJA}) since they can be \emph{global} maximizer or global minimizer of the contrast function.
Let us see the following example for a more detailed investigation.
\begin{exam}\label{exampleSpurious}
Consider the case where two source signals $s_1,s_2$ having identical distribution Bimod(-0.4, 2), a mixing matrix $\A=\I$ and the
contrast function based on the polynomial nonlinearity function $g(x)=x^5$.
We hope to use this particular scenario to illustrate the fixed points of the algorithm that belong to different categories: $\Dds$ (demixing vectors), $\Lds\backslash\Dds$ (spurious attractive fixed points) and $\Fds\backslash\Lds$ (unattractive fixed points).

To find the locations of fixed points, we rely on Lemma \ref{lemmaFds} which states that a vector is a fixed point if and only if it is a zero of $\|\fii(\w)\|$. We plotted in the upper part of Fig. \ref{wei1} the curve of $\|\fii(\w(\theta))\|$ versus $\theta$
using the angular parametrization $\w(\theta)=(\cos(\theta), \sin(\theta))^\TR$.
From the figure, we observe that in the interval $[0, \pi]$ there exist a total of seven zeros, corresponding to seven fixed points. Clearly, among them
$0, \pi/2$ and $\pi$ are demixing vectors.

The rest of those fixed points, namely
$\theta_1=0.089$, $\theta_2=\pi/4$, $\theta_3=1.482$ and $\theta_4=3\pi/4$ do not correspond to any solution of ICA. They can be either attractive or unattractive. To find the attractiveness of these points, we plotted the curve of $\|\f'(\w(\theta))\|$  in the lower part of Fig. \ref{wei1}. According to the figure, the value of $\|\f'(\cdot)\|$ at $\theta_2=\pi/4$ is below the level $\|\f'(\cdot)\|=1$ (the horizonal dash line), while it is above this level at
$\theta_1=0.089$ and $\theta_3=1.482$. This means that $\theta_2$ is a spurious attractive fixed point whereas the other two are unattractive.

To investigate the behavior of FastICA under the presence of these fixed points,
  we plotted in Fig. \ref{wei2} the curve of the contrast function and its values at each algorithm iteration with 100 initial iterates distributing uniformly in $[0, \pi]$. First, we observe that the global maximum of the contrast function is reached at $\theta_2=\pi/4$. This means that searching the global extremum of the function having form of (\ref{contrast1}) may yield a spurious solution.
  Besides, we observe that for any initial iterate between
$\theta_1=0.089$ and $\theta_3=1.482$ (and even for many others outside of this region), the value of the contrast function is updated towards its global maximum, implying that the algorithm will eventually converge to the spurious solution $\theta_2=\pi/4$.
\end{exam}

Although ``kurtosis'' possesses the desired property $\Dds=\Lds$ by Theorem \ref{kurtosis}, we wish to find other nonlinearities that, at least experimentally, share the same property.
In addition to the three classical nonlinearities used in Table \ref{table1}, we have also tested several others such as $g(x)=x^5$ and $g(x)=x^7$, with various source distributions. We found out that for none of them there holds strictly $\Dds=\Lds$. These nonlinearities fail mostly when certain asymmetrical bimodal distribution is involved.

\subsection{Impact of sampling error}
We have just investigated the behaviour of the spurious solutions of FastICA in an somewhat ideal situation where a sufficiently large sample (e.g. $N=5000$) is available.
In this case, the sampling error is negligible and the algorithm behaves as what we anticipate based on the theoretical analysis.
However, when the sample size is small, which is common in practical applications, the estimation error of the FastICA function may be significant.
For a particular realization $\s(1),\ldots,\s(N)$ of the source signal, the sampling error may lead to the following possible situations:
\begin{itemize}
\item[-] FastICA successfully converges to $\hat{\a}_i$, the correct estimator of $\a_i$, but the estimation error $\|\hat{\a}_i-\a_i\|$ is so large that the estimate is no better than a plain guess. 
\item[-] Some theoretically unattractive fixed point becomes attractive with a relatively large convergence domain, constantly absorbing the algorithm to the wrong limit. 
\item[-] Unexpected fixed point emerges, which should not have existed at that place. It locally traps the algorithm in its neighbourhood, if the initial iterate falls within.
\end{itemize}
The behaviour of the algorithm for a particular observation is in general unpredictable, and it is difficult to tell whether a ``bad'' estimate given by FastICA is actually due to an intrinsic spurious solution or due to the effect of sampling error.
In the following example,  we tackle this problem by counting comparing the occurrence of ``bad'' estimates among 10000 independent trials for many different sample sizes.
\begin{exam}\label{example1122a}
 Consider three different scenarios, each involving a different combination of source signals: in Table \ref{table1116a} (a), all five source signals have different distributions and only one of them is Bimod(2,-0.4); in Table \ref{table1116a} (b), two sources are Bimod(2,-0.4); in Table \ref{table1116a} (c), three sources are Bimod(2,-0.4). For each combination of source signals, we tested many different sample sizes (from $N=100$ to $N=10000$) in order to reveal the impact of sampling error. In this example, we take the algorithm stopping criterion as $\epsilon=10^{-8}$. Besides, we use the following the deviation index $\alpha$
 \be
 \alpha(\w_{\infty})\defeq  1- \max_{i=1,\dots,5}\{ |\w_{\infty}^\TR\a_i| \}
 \ee
and take the threshold $\alpha=0.01$, where $\w_{\infty}$ denotes the output of the algorithm.
If $\w_{\infty}$ deviates too much from all of the theoretical demxing vectors $\pm\a_i$ for $i=1,\ldots, 5$,
 that is, if $\alpha(\w_{\infty})>0.01$,  then it is counted as a ``bad'' estimate.

 First, let us look at the count of ``bad'' estimates in each scenario for the largest sample size $N=10000$. In this case, the sampling error is negligible, hence the algorithm should behave as in the asymptotic regime ($N=\infty$).
The result given in Table \ref{table1116a} (a) exhibits no sign of ``bad'' estimates. It  reveals that only one source signal having asymmetrical bimodal distribution such as Bimod(2,-0.4) does not necessarily introduce spurious solutions.
From Table \ref{table1116a} (b), we observe that with the presence of two Bimod(2,-0.4) sources, spurious solutions emerge for ``Gauss'' and ``tanh''. This result is actually expected in view of Proposition \ref{118c} and Table \ref{table1}. In Table \ref{table1116a} (c), there are three sources having Bimod(2,-0.4) rather than two. In this case, more spurious solutions are detected for ``Gauss'' and ``tanh'', compared with Table \ref{table1116a} (b).

The simulation results also show that for all three nonlinearities,
the occurrence of ``bad'' estimates diminishes as the sample size increases.
In particular, given a sufficiently large sample ($N\geq 5000$) and a sufficiently strict stopping criterion ($\epsilon=10^{-8}$), the ``kurtosis'' nonlinearity is almost immune to spurious solutions. For a medium sample size ($500\leq N\leq 1500$), ``kurtosis'' may yield spurious solutions but the occurrence rate is much lower than that of ``Gauss'' and ``tanh''. For a small sample size ($N< 500$), however, the sampling error becomes significant and ``kurtosis'' nonlinearity exhibits no superiority
 compared with the other two competitors in terms of  occurrence rate of spurious solutions.
\end{exam}


\subsection{Practical consideration on the choice of nonlinearity}
The choice of nonlinearity not only determines the estimation performance of the FastICA algorithm, but also has a strong influence on the occurrence of spurious solutions, as shown previously. Therefore, when choosing the nonlinearity function for the application, both aspects should be taken into account.
In what follows, we show how one could avoid spurious solutions without compromising the estimation performance by selecting the nonlinearities wisely.

Using FastICA to estimate optimally the demixing vectors and the source signals consists of three steps \cite{TICHOJA2}:
\begin{enumerate}
\item Run FastICA with a predetermined nonlinearity to obtain an initial estimate of $\a_i$ and $s_i$ for each $i$;
\item Estimate the PDF of $s_i$ based on the previous estimate $\hat{s}_i$, then find the optimal nonlinearity $g_{opt}$ of $s_i$;
\item Run FastICA again with the optimal nonlinearity $g_{opt}$ and the initial iterate $\hat{\a}_i$ obtained in step 1.
\end{enumerate}
Obviously, if the initial estimate $\hat{\a}_i$ obtained during the first step is a spurious solution, then the final result will very likely be incorrect. For this reason, it is of vital importance to have a reliable (even if suboptimal) estimate of the demixing vector in the first place.

To the best of our knowledge, ``kurtosis'' is the only commonly used nonlinearity function that is theoretically free of spurious attractive fixed point. Hence when
 there is no prior information about the distributions of the sources and a sufficiently large sample is available (e, g. $N\geq 1500$ for the case of $d=5$),
 the ``kurtosis'' nonlinearity should be the preferred choice to fulfill step 1.

 By contrast, although ``tanh'' and ``Gauss'' are claimed to be superior to ``kurtosis'' in terms of robustness \cite{HYVABOOK}, these two nonlinearities perform poorly with high occurrence rate of spurious solutions when asymmetrical bimodal distributions Bimod(2,-0.4) are involved, according to Table \ref{table1} and \ref{table1116a}.

Once we successfully localize the source signals using ``kurtosis'' during the first run, we can then proceed with step 2 and step 3.
It is well known that \cite{TICHOJA}
if all the source signals have the same probability density function $p(x)$, then the optimal nonlinearity is the so-called \emph{score function}  :
\be
g_{opt}(x)=-\frac{p'(x)}{p(x)}.
\ee
Using our prior estimate $\hat{s}_i$, we can estimate the PDF of the sources then the score function $\hat{g}_{opt}$. Running FastICA  again with $\hat{g}_{opt}$ will yield an estimator that is asymptotically efficient. The point here is to initiate the algorithm from previously obtained estimates $\hat{\a}_i$ for each $i$.
Since these are reliable estimates of true demixing vectors
 and all demixing vectors are attractive fixed points regardless of the nonlinearity used, running FastICA again with $\hat{g}_{opt}$ and initial iterates $\hat{\a}_i$ will give an  optimal estimator of $\a_i$ without the risk of spurious solutions.
If the source signals are not i.i.d., then we have to estimate separately the optimal nonlinearity for each source and run FastICA multiple times\footnote{One need to employ the symmetrical version of FastICA though.} to achieve the optimal performance. We refer the readers to \cite{TICHOJA2} for more details.

The message here is that ``kurtosis'' is a universal and reliable nonlinearity that is suitable for first run of FastICA.


\section{Conclusion}
In this work, a relationship between the sets $\Dds,\Lds,\Ods$ and $\Fds$ is discovered and the spurious solutions of FastICA are investigated. In the first part of the paper, we established the inclusion
$\Dds\subset\TF\subset\Ods\subset\Fds$ for a general nonlinearity and $\Dds=\Lds\subset\Ods\subset\Fds$ for ``kurtosis''. In the second part, we showed
 that unattractive fixed points widely exist on the sphere regardless of the nonlinearity function and the source distributions involved. These unattractive fixed points may lead to the occurrence of spurious solutions if the initial iterate of the algorithm falls within some small neighborhood of one of those points. 
We showed that this type of spurious solution, being already statistically rare, can be further reduced
by adopting 
a tight stopping criterion such as $\epsilon=10^{-8}$.  Another category of spurious solutions consists of attractive fixed points of FastICA. This type of spurious solutions are present when certain bimodal distributions with Gaussian mixtures are involved. In this case,  common nonlinearities such as ``Gauss'' and ``tanh'' will fail and only ``kurtosis'' may give reliable results. For this reason, ``kurtosis'' nonlinearity is our recommended choice for the initial run of FastICA.

\section{Appendix}
\section{Proof of Theorem \ref{kurtosis}\label{proofKurtosis}}
Without loss of generality, we may suppose that the mixing matrix $\A$ is an identity matrix. Let $\v$ be a fixed point, and write $\v=(v_1,\ldots,v_d)^\TR$. By Lemma (\ref{lemmaFds}), we have for any $i\in\{1,\ldots,d\}$,
\be
\Eb\Big[\sum_{j=1}^d({v_j s_j})^3s_i\Big] & = &  \Eb\Big[\sum_{j=1}^d({v_j s_j})^4\Big]v_i,
\ee
or equivalently, after some algebraic simplifications,
\ben\label{619a}
\kappa_i v_i^3 + 3 v_i & = & \Big(\sum_{j=1}^d v_j^4 \kappa_j +3\Big)v_i,
\een
where $\kappa_i\defeq \Eb[s_i^4]-3\neq 0$ by assumption $\Ac_2$ (see Section \ref{assumption12}).
Denote by $\Ic$ the set of indices $i$ such that $v_i\neq 0$.
It follows from (\ref{619a}) that for $i\in\Ic$
\ben
 v_i^2 & = & \frac{1}{\kappa_i}\sum_{j=1}^d v_j^4 \kappa_j.  \label{619b}
\een
 Since $\sum_{i=1}^d v_i^2=1$, we deduce from (\ref{619b})
\be
\sum_{i\in\Ic} \Big( \frac{1}{\kappa_i}\sum_{j=1}^d v_j^4 \kappa_j  \Big) &= &1,
\ee
or equivalently
\be
\sum_{j=1}^d v_j^4 \kappa_j&= & \Big(\sum_{i\in\Ic} \kappa_i^{-1} \Big)^{-1}.
\ee
Then we can rewrite (\ref{619a}) as
\be
v_i^2 = \kappa_i^{-1} \Big(\sum_{j\in\Ic} \kappa_j^{-1} \Big)^{-1},\quad i\in\Ic.
\ee
Now let us calculate $\M\defeq \Eb[g'(\v^\TR\s)\s\s^\TR]$. We have
\be
\M_{ij}&=&\Eb\Big[3 \sum_k (v_k s_k)^2 s_is_j \Big] = 6v_iv_j\\
\M_{ii}&=&\Eb\Big[3 \sum_k (v_k s_k)^2 s_i^2 \Big]=3(\kappa_i v_i^2 + 2v_i^2 + 1).
\ee
From this, we deduce that $\M=3(2\v\v^\TR + \I +\D)$, where $\D$ is a diagonal matrix with the $i$th diagonal entry
$\D_i=\kappa_i v_i^2$.  Since
\be
\alpha(\v)=\Eb\Big[3- \Big(\sum_{i=1}^dv_is_i\Big)^4\Big]=-\sum_i v_i^4 \kappa_i,
\ee
it follows from (\ref{619b}) that $\D_i=-\alpha(\v)$ for $i\in\Ic$ and $\D_i=0$ otherwise.
Besides, since $\Eb[g'(\v^\TR\s)\I]=3\I$, we have
$\Eb[g'(\v^\TR\s)(\I-\s\s^\TR)]=3\I-\M=-3(\D + 2\v\v^\TR)$.
From this we deduce that
\ben
\f'(\v)&=&\frac{(\I-\v\v^\TR)\Eb[g'(\v^{\TR}\x)(\I-\x\x^{\TR})]}{ |\alpha(\v) |}  \nonumber \\
&=&\frac{-3(\I-\v\v^\TR)(\D+2\v\v^\TR)}{ |\alpha(\v) |} \nonumber \\
&=& 3\sign(\alpha(\v)) (\I-\v\v^\TR)\bar{\D}, \label{619c}
\een
where $\bar{\D}\defeq -\D/|\alpha(\v)|$. Clearly, the diagonal entry of $\bar{\D}$ satisfies $\bar{\D}_i=1$ for $i\in\Ic$ and $\bar{\D}_i=0$ otherwise. Besides, it is easy to see that $\bar{\D}\v=\v$, which implies $\spanv(\v)\subset \range(\bar{\D})$.
Denote by $\#\Ic$ the cardinal of $\Ic$.
Since $\dim(\spanv(\v))=1$ and $\dim(\range(\bar{\D}))=\#\Ic\geq 1$, this inclusion becomes an equality if and only if
$\#\Ic=1$, i.e. there is exactly one entry $v_i\neq 0$, or equivalently, $\v=\e_i\in\Dds$ for some $i$. If this is the case, then we have immediately $\f'(\v)=0$ by (\ref{619c}). Otherwise, take any vector $\u=(u_1,\ldots,u_d)^\TR$ such that $\|\u\|=1$, $\u^\TR\v=0$ and $u_i=0$ for $i\neq \Ic$. We have
\be
\f'(\v)\u=3\sign(\alpha(\v)) (\I-\v\v^\TR)\bar{\D}\u=3\sign(\alpha(\v))\u.
\ee
 On the one hand, by the submultiplicativity of spectral norm,
\be
\|\f'(\v)\|\leq 3\|\I-\v\v^\TR\|  \|\bar{\D}\|=3;
\ee
on the other hand, there also holds
\be
\|\f'(\v)\|=\sup_{\w\in\Sc}\|\f'(\v)\w\|\geq \|\f'(\v)\u\|=3.
\ee
It then follows that $\|\f'(\v)\|=3$ for any $\v\in \Fds\backslash\Dds$. This fact also implies $\Dds=\Lds$.
\section{Proof of Proposition \ref{main2}  \label{proofmain2}}
Without loss of generality, in what follows we take $\A=\I$ for simplicity.
\subsection{Case $d=2$}
Let us first consider the simplest case $d=2$. If $\alpha(\e_1)$ and $\alpha(\e_2)$ have the same sign, say, positive, then $\e_1$ and $\e_2$ are local minimizers of the contrast function $\Jc(\cdot)$ on $\Sc$. Write
\ben
\w(\theta)&=& \cos(\theta)\e_1+ \sin(\theta)\e_2 =  \begin{pmatrix}\cos(\theta) \\ \sin(\theta)\end{pmatrix},  \label{1123a}\\
f(\theta) &=&\Jc(\w(\theta))=\Eb[G(\cos(\theta)s_1 + \sin(\theta)s_2)]  \label{614b}.
\een
Then it is easy to see that $\theta_1=0$ and $\theta_2=\pi/2$ are local minimizers of $f(\theta)$ on $\Rb$.
From that we deduce immediately that $f$ reaches its local maximum at some internal point $\theta_0\in (\theta_1,\theta_2)$.
The corresponding vector
$\v\defeq \big(\cos(\theta_0), \sin(\theta_0)\big)^\TR$
 is then a local maximizer of $\Jc(\w)$ on $\Sc$.

We actually proved the following result:
\begin{lemm}\label{lemma614}
Let $s_1$ and $s_2$ be two random variables such that the quantity
$\Eb[g'(s_i) - g(s_i)s_i]$
has the same sign for $i=1,2.$ Then
$f(\theta)\defeq \Eb[G(\cos(\theta)s_1 + \sin(\theta)s_2)]$
has a local optimum at some $\theta_0\in (0,2\pi)$. Moreover, the angle $\theta_0$ satisfies
\ben
&&\Eb\Big[g\big(\w(\theta_0)^{\TR}\s\big)\s\Big] =  \Eb\Big[g\big(\w(\theta_0)^{\TR}\s\big)\w(\theta_0)^\TR\s\Big]\w(\theta_0) \label{eq1}
 \een
\end{lemm}
 Equality (\ref{eq1}) come directly from Lemma \ref{lemmaFds}.
\subsection{Case $d>2$}
Suppose $\e_i$ and $\e_j$ are two demixing vectors such that both $\alpha(\e_i)$ and $\alpha(\e_j)$ are positive.
Write
\be
f^{(i,j)}(\theta) &\defeq& \Eb[G(\cos(\theta)s_i + \sin(\theta)s_j)].
\ee
By Lemma \ref{lemma614}, there exists $\theta'\in(0,2\pi)$ such that $\theta'$ maximizes $f^{(i,j)}(\theta)$ and satisfies
\ben
\Eb\Big[g\big(\w(\theta')^{\TR}\s_{ij}\big)\s_{ij}\Big]\!\!=\!  \Eb\Big[g\big(\w(\theta')^{\TR}\s_{ij}\big)\w(\theta')^\TR\s_{ij}\Big]\w(\theta'),\label{eq3}
\IEEEeqnarraynumspace
\een
where $\s_{ij}=(s_i,s_j)^\TR$.
Consider vector
$\u=(u_1,\ldots,u_d)$ with
$u_i=\cos(\theta')$, $u_j=\sin(\theta')$ and $u_k=0$ for $k\neq i,j$. Clearly,
 $\u^\TR\s=\cos(\theta')s_i + \sin(\theta')s_j=\w(\theta')^{\TR}\s_{ij} $. It then follows from (\ref{eq3}) that
 \ben \label{eq5}
 \Eb[g(\u^{\TR}\s)\s] =  \Eb[g(\u^{\TR}\s)\u^\TR\s]\u,
 \een
  which implies $\v\in\Fds$ by Lemma \ref{lemmaFds}.

In the particular case that $s_1$ and $s_2$ have the same distribution, we must have
$\cos(\theta')=\sin(\theta')=1/\sqrt{2}$ by symmetry. This means $\theta'=\pi/4$.
\section{Proof of Proposition \ref{118c}  \label{ProofSpurious}}
Suppose that $\u\in\Rb^2$ is a spurious attractive fixed point in the case $d=2$ with $s_1\sim \Dc_1$ and $s_2\sim \Dc_2$ . Then $\|\f'(\u)\|<1$ and
\be
\u^\TR\x= cs_1 + \sqrt{1-c^2}s_2
\ee
 for some real scalar $c\in (0,1)$. Now let us consider the case $d=n>2$ with $s_i\sim \Dc_1$ and $s_j\sim\Dc_2$ for some indices $i\neq j$.
 In the sequel, we assume  $i=1$, $j=2$ for simplicity.
Take
\be
\v=c\a_1 + \sqrt{1-c^2}\a_2\in\Rb^d.
\ee
It is easy to see that  $\v^\TR\x=cs_1 + \sqrt{1-c^2}s_2$ and $\v\in\Fds$. Next, we show $\|\f'(\v)\|=\|\f'(\u)\|<1$, where $\f'(\v)$ and $\f'(\u)$ are respectively $n\times n$ and $2\times 2$ matrices. Note that these two ``$\f'(\cdot)$'' are different mappings for they are determined by different ICA models.
Denote respectively by $\A_\u$ and $\A_\v$ the mixing matrices for each case.
  Since the mixing matrix $\A_\u$ is orthogonal, for any $\w\in\Rb^2$ we have
\ben
&&\|(\I-\w\w^\TR)\Eb[g'(\w^{\TR}\x)(\I-\x\x^{\TR})]\|   \nonumber \\
&=&\|(\I-\A_\u^\TR\w\w^\TR\A_\u)\Eb[g'(\w^{\TR}\A_\u\x)(\I-\s\s^{\TR})]\|. \label{1114a}
\een
Similar equality also holds for $\A_\v$ and $\w\in\Rb^n$.
Denote
\be
\B_\u & \defeq & (\I-\A_\u^\TR\u\u^\TR\A_\u)\Eb[g'(\u^{\TR}\A_\u\s)(\I-\s\s^{\TR})] \\
\B_\v & \defeq & (\I-\A_\v^\TR\v\v^\TR\A_\v)\Eb[g'(\v^{\TR}\A_\v\s)(\I-\s\s^{\TR})].
\ee
Using (\ref{1114a}) and (\ref{fprime}), we get
\ben \label{1112a}
\|\f'(\u)\|=\frac{\|\B_\u\|}{|\alpha(\u)|},\quad \|\f'(\v)\|=\frac{\|\B_\v\|}{|\alpha(\v)|}.
\een
Notice that
\be
\u^\TR\A_\u&=&(c,\sqrt{1-c^2})\in\Rb^2 \\
\v^\TR\A_\v&=&(c,\sqrt{1-c^2},0,\ldots,0)\in\Rb^d
\ee
by the construction of $\u$ and $\v$.
This implies $\u^\TR\x$ and $\v^\TR\x$ have the same distribution and therefore $\alpha(\u)=\alpha(\v)$.
Besides, it is easily seen that
\ben\label{1112}
\B_{\v}&=&\begin{pmatrix} \B_\u & \mathbf{0}  \\ \mathbf{0} & \mathbf{0}  \end{pmatrix}.
\een
From (\ref{1112}), we can deduce that $\|\B_\u\|=\|\B_\v\|$. Finally, combining this result with (\ref{1112a}) gives
$\|\f'(\v)\|<1$.

\section{Probability distributions used in Table \ref{table1}  \label{PDF}}
\subsection{Generalized Gaussian distribution}
The generalized Gaussian density function with parameter $\alpha$, zero mean and unit variance is given by
\be
f_{\alpha}(x)=\frac{\alpha\beta_{\alpha}}{2\Gamma(1/\alpha)}\exp{\{-(\beta_{\alpha}|x|)^{\alpha}\}},
\ee
where $\alpha$ is a positive parameter that controls the distribution¡¯s exponential rate of decay, $\Gamma$ is the Gamma function, and
\be
\beta_{\alpha}=\sqrt{\frac{\Gamma(3/\alpha)}{\Gamma(1/\alpha)}}.
\ee
This generalized Gaussian family encompasses the ordinary
standard normal distribution for $\alpha=2$ , the Laplace distribution for $\alpha=1$, and the uniform distribution in the limit $\alpha\to\infty$.

\subsection{Bimodal distribution with Gaussian mixture}
\begin{figure}[!t]
\centerline{
{\includegraphics[width=3.5in]{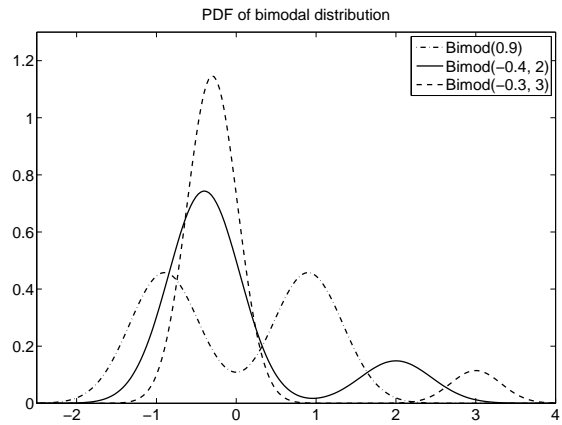}} }
\caption{PDF curves of three bimodal distributions used in Table \ref{table1}. \label{wei5}}
\end{figure}

The bimodal distribution used in this paper consists of a mixture of two Gaussian distribution. Define random variable
\be
X=Z Y_1 + (1-Z)Y_2,
\ee where
$Y_i\sim\Nc(\mu_i,\sigma_i^2)$ for $i=1,2$ and $Z\sim\Bc(p)$. Here, $\Bc(p)$ denotes the Bernoulli distribution. If we impose that $\sigma^2_1=\sigma^2_2=\sigma^2$ and that $X$ have zero mean and unit variance, then it is easy to obtain the following relationship:
\be
p =\frac{|\mu_2|}{|\mu_1|+|\mu_2|}, \quad \sigma^2  =  1-|\mu_1\mu_2|,
\ee
where $|\mu_1|, |\mu_2|\leq 1$. Since the distribution of $X$ is completely determined by $\mu_1,\mu_2$, we take them as controlling parameter and denote by
 ``Bimod$(\mu_1,\mu_2)$'' the distribution of $X$. The PDF of
Bimod$(\mu_1,\mu_2)$ can be given explicitly (see Fig. \ref{wei5}):
\be
f_X(x)= p f_{Y_1}(x) + (1-p)f_{Y_2}(x),
\ee
where $f_{Y_i}$ is the PDF of $Y_i\sim\Nc(\mu_i,\sigma^2)$ for $i=1,2$. Note that if $\mu_1=-\mu_2$, then $p=1/2$ and the distribution of $X$ becomes symmetrical. In this case, we write simply ``Bimod($\mu$)'' with $\mu=|\mu_1|$. Note also that the ``bpsk'' distribution is actually Bimod(1).

\section*{Acknowledgement}
The author would like to express the deepest gratitude to Prof. A. Dermoune for his invaluable guidance.
The author would also like to thank the anonymous referees
for carefully reading the manuscript and for giving us many
helpful and constructive suggestions resulting in the present
work.

\bibliographystyle{IEEEtran}
\bibliography{IEEEabrv,IeeeBib}

  \begin{table*}
  \caption{Total number of spurious solutions obtained among 10000 independent trials with random initial iterate; The source signals have identical distribution with $N=5000$.  \label{table1}}
\centerline
{
\begin{tabular}{|l||ccccc|ccccc|ccccc|ccccc|}
\hline
 \multicolumn{1}{|c}{$d=2$}  &  \multicolumn{5}{c}{Gauss}         &  \multicolumn{5}{c}{Tanh}           &  \multicolumn{5}{c|}{Kurtosis}   \\ \hline
 PDF                    &$\|\f'\|$&  $10^{-4}$ & $10^{-6}$ & $10^{-8}$ & $\times10$
                        &$\|\f'\|$&  $10^{-4}$ & $10^{-6}$ & $10^{-8}$ & $\times10$
                        &$\|\f'\|$&  $10^{-4}$ & $10^{-6}$ & $10^{-8}$ & $\times10$ \\ \hline
Sinus                   &6.48& 43&0&0      & 0                    &5.93&24&0&0& 0                            &3&103&4&0    &0                         \\
Uniform                 &5.12& 31&0& 0 &  0                       &4.68&16&0&0& 0                            &3&96&7&0&  0                            \\
GG(3)                   &3.92& 24&10&2  & 0                       &3.71&47&6&0& 0                            &3&95&8&0 &  0                          \\
Laplace                 &2.26& 106&7&1 & 0                        &2.41&130&24&0& 0                          &3&96&8&1 & 0                           \\
GG(0.5)                 &1.55& 312&19&1 & 2                       &1.70&250&17&6& 0                         &3&105&15&3& 0                            \\
Bimod(0.9)              &6.05&46&4&0  & 0                         &5.65&33&5&0& 0                            &3&80&7&0  & 0                          \\
Bpsk                    &13.2&15&0&0 &  0                         &17.3&5&0&0&  0                            &3&84&9&0  & 0                      \\ \hline\hline
Bimod(-0.4, 2)          &0.78& 3218&2625&2236&2518                &0.90&2402&1365&1025& 1147                     &3&95&14&0 & 1                   \\
Bimod(-0.3, 3)          &0.97& 1252& 848&707 &816                 &1.24&428&94&2&55                       &3&77&8&0  &0                           \\ \hline
\end{tabular}}
\vspace{0.5cm}
\centerline
{
\begin{tabular}{|l||ccccc|ccccc|ccccc|ccccc|}
\hline
 \multicolumn{1}{|c}{$d=3$}  &  \multicolumn{5}{c}{Gauss}         &  \multicolumn{5}{c}{Tanh}           &  \multicolumn{5}{c|}{Kurtosis}   \\ \hline
 PDF                    &$\|\f'\|$&  $10^{-4}$ & $10^{-6}$ & $10^{-8}$ & $\times10$
                        &$\|\f'\|$&  $10^{-4}$ & $10^{-6}$ & $10^{-8}$ & $\times10$
                        &$\|\f'\|$&  $10^{-4}$ & $10^{-6}$ & $10^{-8}$ & $\times10$ \\ \hline
Sinus                   &6.48 &11 &0 & 0& 0                         &5.93&4&0&0&0                             &3&28&0&0&0                         \\
Uniform                 &5.12 &17 &0 & 0& 0                         &4.68&3&0&0&0                             &3&43&0&0&0                         \\
GG(3)                   &3.92 &24 &0 & 0& 0                         &3.71&20&0&0&0                            &3&30&0&0&0                         \\
Laplace                 &2.26 &68 &4 & 0&  0                        &2.41&52&7&1& 1                           &3&31&0&0&0                         \\
GG(0.5)                 &1.55 &333 &25 &0 & 14                      &1.70&191&10&0&1                          &3&20&0&0&0                         \\
Bimod(0.9)              &6.05 &16 &0 & 0 & 0                        &5.65&8&0&0&0                             &3&47&6&0&0                         \\
Bpsk                    &13.2 &0 &0 & 0 & 0                         &17.3&4&0&0&0                             &3&51&5&0&0                       \\ \hline\hline
Bimod(-0.4, 2)          &0.78 &3572 & 3261 & 3029 &3291             &0.90&2092&1723&1343&1683                 &3&52&10&1&1                        \\
Bimod(-0.3, 3)          &0.97 &209 &144 & 132 & 180                 &1.24&22&0&0&0                            &3&13&0&0&0                           \\ \hline
\end{tabular}}
\vspace{0.5cm}
\centerline
{
\begin{tabular}{|l||ccccc|ccccc|ccccc|ccccc|}
\hline
 \multicolumn{1}{|c}{$d=5$}  &  \multicolumn{5}{c}{Gauss}         &  \multicolumn{5}{c}{Tanh}           &  \multicolumn{5}{c|}{Kurtosis}   \\ \hline
 PDF                    &$\|\f'\|$&  $10^{-4}$ & $10^{-6}$ & $10^{-8}$ & $\times10$
                        &$\|\f'\|$&  $10^{-4}$ & $10^{-6}$ & $10^{-8}$ & $\times10$
                        &$\|\f'\|$&  $10^{-4}$ & $10^{-6}$ & $10^{-8}$ & $\times10$ \\ \hline
Sinus                   &6.48 &0 &0 &0 &0                          &5.93&0&0&0&0                             &3&10&0&0&0                         \\
Uniform                 &5.12 &3 &0 &0 &0                          &4.68&1&0&0&0                             &3&19&0&0&0                          \\
GG(3)                   &3.92 &13 &0 &0 &0                         &3.71&10&0&0&0                            &3&14&0&0&0                         \\
Laplace                 &2.26 &51 &0 &0 &0                         &2.41&45&4&0&0                            &3&23&0&0&0                       \\
GG(0.5)                 &1.55 &178 &17 &3 &35                      &1.70&108&13&1&1                          &3&18&0&0&0                      \\
Bimod(0.9)              &6.05 &1 &0 &0 &0                          &5.65&2&0&0&0                             &3&20&0&0&0                      \\
Bpsk                    &13.2 &0 &0 &0 &0                          &17.3&1&0&0&0                             &3&25&4&0&0                       \\ \hline\hline
Bimod(-0.4, 2)          &0.78 &5502 &5330 &5115 &5319              &0.90&3086&2752&2577&2820                 &3&44&10&5&17                      \\
Bimod(-0.3, 3)          &0.97 &1033 &707 &528 & 824                &1.24&166&9&5&59                          &3&12&0&0& 0                        \\ \hline
\end{tabular}}
\end{table*}

\newpage

\newpage
\begin{table}
  \caption{Total number of ``bad'' estimates obtained among 10000 independent trials with random initial iterate. \label{table1116a}}
  \centering
  {
  \begin{tabular}{c}
  (a) Five source signals have different distributions
  \end{tabular}
  }
  \centering
{
\begin{tabular}{c|c|c|c|c}
\hline
 $s_1$  & $s_2$     & $s_3$      & $s_4$          & $s_5$    \\
Uniform & Laplace   & GG(2)      & GG(3)          & Bimod(2,-0.4)   \\ \hline
\end{tabular}
}
\vspace{0.2cm}
\centering
{
\begin{tabular}{l|cccccc}
\hline
 Sample size  & 100 & 200      & 500  & 1500 & 5000 & 10000   \\ \hline
Gauss          & 1002 & 318        &69    &5     & 0 & 0                                  \\
tanh           & 785 & 253        &41    &8     & 0  & 0                               \\
kurtosis       & 957 & 270        &54    &11     &0  & 0                             \\ \hline
\end{tabular}
}
\vspace{0.5cm}
  \centering
  {
  \begin{tabular}{c}
  (b) $s_4,s_5\sim$Bimod(2,-0.4)
  \end{tabular}
  }
   \centering
{
\begin{tabular}{c|c|c|c}
\hline
 $s_1$  & $s_2$     & $s_3$      & $s_4,s_5$    \\
Uniform & Laplace   & GG(3)      & Bimod(2,-0.4)    \\ \hline
\end{tabular}
}
\vspace{0.2cm}
\centering
{
\begin{tabular}{l|cccccc}
\hline
 Sample size   & 100 & 200      & 500  & 1500 & 5000 & 10000   \\ \hline
Gauss          & 2066 & 675     & 251  & 165  &139   &  156                                \\
tanh           & 1823 & 701     & 198  & 54   & 55   &  65                              \\
kurtosis       & 1840 & 792     & 204  & 25   & 0    &  0                         \\ \hline
\end{tabular}}
\vspace{0.5cm}
  \centering
  {
  \begin{tabular}{c}
  (c) $s_3,s_4,s_5\sim$Bimod(2,-0.4)
  \end{tabular}
  }
   \centering
{
\begin{tabular}{c|c|c}
\hline
 $s_1$  & $s_2$     & $s_3,s_4,s_5$    \\
Uniform & Laplace   & Bimod(2,-0.4)   \\ \hline
\end{tabular}
}
\vspace{0.2cm}
\centering
{
\begin{tabular}{l|cccccc}
\hline
 Sample size   & 100 & 200      & 500  & 1500 & 5000 & 10000   \\ \hline
Gauss          & 2084 & 931     & 555  & 516  & 708  & 783                                \\
tanh           & 1801 & 680     & 236  & 178  & 228  & 258                              \\
kurtosis       & 1769 & 707     & 324  & 81   & 1    &  0                         \\ \hline
\end{tabular}}
\end{table}
\newpage
\end{document}